\pdfoutput=1

\documentclass[11pt]{article}

\usepackage[final]{acl}

\usepackage{times}
\usepackage{latexsym}

\usepackage[T1]{fontenc}

\usepackage[utf8]{inputenc}

\usepackage{microtype}

\usepackage{inconsolata}

\usepackage{algorithm}
\usepackage{graphicx}
\usepackage{booktabs} 

\newcount\Comments  
\newcommand{\kibitz}[2]{\ifnum\Comments=1{\color{#1}{#2}}\fi}

\definecolor{darkgreen}{rgb}{0.3,0.7,0.1}

\usepackage{dsfont}
\usepackage{soul}
\usepackage{url}
\usepackage{float}
\usepackage[utf8]{inputenc}
\usepackage{graphicx,multirow,multicol}
\usepackage{amsmath}
\usepackage{epstopdf}
\usepackage{amssymb}
\usepackage{caption}
\usepackage{pifont}         
\usepackage{booktabs,bbm,dsfont}
\usepackage{enumitem}
\usepackage{booktabs,wrapfig} 
\usepackage{graphicx,array,multirow}
\usepackage{color,soul,xspace}
\usepackage{xcolor,colortbl}
\usepackage{natbib}
\usepackage{bm}
\usepackage{amsthm}
\usepackage{amsfonts}
\usepackage{tikz}
\usepackage{subcaption}
\usepackage[noend]{algpseudocode}
\usepackage{algorithm}

\definecolor{Gray}{gray}{0.85}
\definecolor{Yellow}{rgb}{0.5,0.5,0.5}
\usepackage[noend]{algpseudocode}
\algblockdefx[Foreach]{Foreach}{EndForeach}[1]{\textbf{for each} #1 \textbf{do}}{\textbf{end for}}

\algnewcommand{\IIf}[1]{\State\algorithmicif\ #1\ \algorithmicthen}
\algnewcommand{\EndIIf}{\unskip\ \algorithmicend\ \algorithmicif}
\usepackage{comment}
\usepackage{epsfig}

\usepackage{mathrsfs,mdwlist,enumitem}
\usepackage[group-separator={,}]{siunitx}
\usetikzlibrary{positioning}

\newcommand{\cora}{\textsc{Cora}\xspace}
\newcommand{\liar}{\textsc{Liar}\xspace}
\newcommand{\wiki}{\textsc{WikiCS}\xspace}
\newcommand{\amazonprod}{\textsc{Amazon-Product}\xspace}
\newcommand{\amazonprodsub}{\textsc{Amazon}\xspace}

\newcommand{\methodname}{\textsc{Gspell}\xspace}
\newcommand{\methodnameplus}{\textsc{Gspell}${}^+$\xspace}

\newcommand{\gnns}{\textsc{Gnn}s\xspace}
\newcommand{\gnn}{\textsc{Gnn}\xspace}

\setlist{nolistsep,leftmargin=*}

\newtheorem{defn}{\textbf{Definition}}

\definecolor{babypink}{rgb}{0.96, 0.76, 0.76}
\definecolor{top1}{HTML}{a5dc82}
\definecolor{top2}{HTML}{dff3d9}

\definecolor{c1}{HTML}{d5e8d4}
\definecolor{c1_1}{HTML}{82b366}
\definecolor{c2}{HTML}{ffe6cc}
\definecolor{c2_1}{HTML}{d79b01}
\definecolor{c3}{HTML}{dae8fc}
\definecolor{c3_1}{HTML}{6c8ebf}
\definecolor{text_grey}{HTML}{5e5e5e}
\hypersetup{hidelinks}

\usepackage{graphicx} 

\usepackage{xcolor}
\usepackage[most]{tcolorbox}
\usepackage{array}
\usepackage{ragged2e}

\definecolor{yesGreen}{RGB}{49,163,84}
\definecolor{noRed}{RGB}{203,24,29}
\definecolor{catBlue}{RGB}{33,113,181}
\definecolor{embedIndigo}{RGB}{76, 64, 142}  
\definecolor{tagGray}{gray}{0.35}
\definecolor{softGray}{gray}{0.97}
\definecolor{softFrame}{gray}{0.75}

\tcbset{
  tightbox/.style={
    colback=softGray,
    colframe=softFrame,
    boxrule=0.4pt,
    arc=1pt, outer arc=1pt,
    left=3pt,right=3pt,top=3pt,bottom=3pt,
    boxsep=2pt,
    width=\linewidth,
  },
  tighttitle/.style={
    fonttitle=\bfseries,
    title filled=false,
  }
}

\newcommand{\SupportYes}{\textbf{\textcolor{yesGreen}{YES}}}
\newcommand{\SupportNo}{\textbf{\textcolor{noRed}{NO}}}
\newcommand{\PredCat}[1]{\textbf{\textcolor{catBlue}{#1}}}
\newcommand{\EmbedText}[1]{\textcolor{embedIndigo}{\texttt{#1}}}

\newcommand{\BeginTK}{\textcolor{tagGray}{\texttt{\textless\textless BEGIN TARGET KEYWORDS\textgreater\textgreater}}}
\newcommand{\EndTK}{\textcolor{tagGray}{\texttt{\textless\textless END TARGET KEYWORDS\textgreater\textgreater}}}
\newcommand{\NeighborBeginTK}{\textcolor{tagGray}{\texttt{\textless\textless BEGIN KEYWORDS\textgreater\textgreater}}}
\newcommand{\NeighborEndTK}{\textcolor{tagGray}{\texttt{\textless\textless END KEYWORDS\textgreater\textgreater}}}

\title{From Nodes to Narratives: Explaining Graph Neural Networks with LLMs and Graph Context}

\author{Peyman Baghershahi\textsuperscript{1}\thanks{Equal contribution} \quad Gregoire Fournier\textsuperscript{1}\footnotemark[1] \quad Pranav Nyati\textsuperscript{2} \quad Sourav Medya\textsuperscript{1} \\
 \textsuperscript{1}University of Illinois Chicago,\\
 \textsuperscript{2}Indian Institute of Technology Kharagpur\\
 \textsuperscript{1}\texttt{\{pbaghe2, gfourn2, medya\}@uic.edu; \textsuperscript{2}\texttt{pranavnyati26@kgpian.iitkgp.ac.in}}
 }

\begin{document}
\maketitle
\begin{abstract}
Graph Neural Networks (GNNs) have emerged as powerful tools for learning over structured data, including text-attributed graphs (TAGs), which are common in domains such as citation networks, social platforms, and knowledge graphs. GNNs are not inherently interpretable and thus, many explanation methods have been proposed. However, existing explanation methods often struggle to generate interpretable, fine-grained rationales, especially when node attributes include rich natural language. In this work, we introduce \methodname, a lightweight, post-hoc framework that uses large language models (LLMs) to generate faithful and interpretable explanations for GNN predictions. \methodname projects GNN node embeddings into the LLM embedding space and constructs hybrid prompts that interleave soft prompts with textual inputs from the graph structure. This enables the LLM to reason about GNN internal representations and to produce natural-language explanations, along with concise explanation subgraphs. Our experiments across real-world TAG datasets demonstrate that \methodname achieves a favorable trade-off between fidelity and sparsity, while improving human-centric metrics such as insightfulness. \methodname sets a new direction for LLM-based explainability in graph learning by aligning GNN internals with human reasoning.



\end{abstract}

\section{Introduction}

Graph neural networks (GNNs) have witnessed rapid adoption across a wide range of critical real-world applications, including healthcare~\cite{zitnik2018modeling,zitnik2019machine}, drug design~\cite{xiong2021graph,drug-design, drug-ligand}, recommender systems~\cite{chen2022grease}, and fraud detection~\cite{fraud-detect}. The high-stakes nature of these domains demands that the predictions made by GNNs be not only accurate but also trustworthy. One powerful approach to building trust in deep learning models is to provide meaningful explanations that justify their predictions~\cite{trust-AI}. Such explanations, which may highlight important substructures in the input~\cite{pgexplainer, Graph-mask, subgraphX,armgaan2024graphtrail}, or offer counterfactual scenarios~\cite{cfgnnex, cf^2-counter, induce,gcfexplainer,fournier2025comrecgc}. However, generating explanations for GNNs is inherently challenging due to the combinatorial nature of graph data and the joint influence of node attributes and edge connections. These complexities make GNN explainability a non-trivial problem and have led to the development of a wide array of explanation techniques~\cite{surveyonexplainability2023,survey-counter}. 

\textbf{Need for LLM-based GNN explainers.} There has been a growing interest in the integration of GNNs and LLMs to enhance GNN task performance on text-attributed graphs (TAGs). Essentially, these methods leverage LLMs in different architectural orders to 1) attain rich representations of the input TAGs, task descriptions, and task-specific few-shot examples for the GNN predictors \cite{oneforall2024, gaugllm2024}, or 2) reason over the TAGs directly \cite{talklikegraph2024} or along with GNN-enhanced soft prompts to make predictions \cite{llaga2024, graphgpt2024}. However, there is a lack of focus on using LLMs to explain GNN predictions and clarify the reasoning behind these black-box models.

Traditional GNN explainers aim to identify influential subgraphs or feature subsets that contribute to predictions \cite{gnn_explainer, pgexplainer}. While effective for graphs with simple node features, these methods perform poorly on TAGs, where node information is expressed in natural language. Moreover, these explanations are often presented as subgraphs that are not human-interpretable to users. LLMs, on the other hand, excel at reasoning over textual content and can generate coherent, human-interpretable rationales. Thus, their integration into GNN explanation pipelines can improve the interpretability of the GNN outputs on TAGs.

\paragraph{Limitations of Existing LLM-based GNN Explainers.} The existing frameworks (detailed in Sec. \ref{sec:RW_LLM_explain}) employing LLMs for explaining GNN have the following shortcomings:
\textit{(1) Template dependency and alignment complexity: }Aligning the GNN explainer's output with the LLM acceptable input necessitates rigid templates, handcrafted scores, or training/fine-tuning of the GNN explainer. 
\textit{(2) Missed GNN internals: } Existing approaches fail to directly leverage the rich internal representations of the GNN, making the explanations generic or unfaithful to the internal working of the GNNs. 
\textit{(3) Bias from suboptimal explainers: } Invoking an external GNN explainer can bias the reasoning and judgment of the LLM. This is critical when the GNN explainers are suboptimal, and they signal noisy information to LLM, while the LLM can potentially better infer the pivotal causes of why the GNN has made specific predictions, particularly for TAGs where the textual attributes carry rich information about the graph entities. 

\paragraph{Our Contributions.}In this work, we present \methodname (\underline{G}NN \underline{S}oft \underline{P}rompted \underline{E}xplanation with \underline{LL}Ms), a lightweight, post-hoc explanation framework for TAGs that integrates the representational power of GNNs with the reasoning capability of LLMs. Our contributions are as follows:

\begin{itemize}
\item \textit{Novel Method}: We introduce a method that bypasses traditional GNN explainer modules by using LLMs as direct interpreters of model behavior. \methodname generates fine-grained, natural language explanations (NLEs) unreliant on handcrafted templates or perturbation-based saliency masks, reducing external explainers' bias.
\item \textit{Soft Prompt Integration}: \methodname aligns the internal representations of a trained GNN with the token-level embedding space of an LLM via a novel embedding projector. This projection enables GNN features to be treated as soft prompts in a hybrid prompt that blends structural and textual cues, allowing the LLM to reason over the graph’s latent space.
\item \textit{Interpretable Explanation Subgraphs}: Unlike prior methods that return subgraph masks, \methodname produces interpretable NLE subgraphs of node-level support decisions.
\item \textit{Training-Agnostic and Plug-and-Play Deployment}: \methodname does not require fine-tuning the GNN or the LLM, making it easily deployable on pre-trained models without retraining. This makes it adaptable for real-world applications.

\end{itemize}

\section{Related Work}
\label{sec:RW_LLM_explain}

We describe the related studies on the integrated LLM and GNN explainer methods. The explanation capability of LLMs has been applied to graph problems in task-oriented contexts \cite{grefer2025, xrec2024}. There have been attempts to guide GNNs using explanations generated by LLMs; for example, in \cite{tape2024}, an LLM generates node predictions and textual explanations of the reasoning behind them.

\textbf{\textit{LLM explanation to enhance explainers. }}LLMs have been utilized to enhance the explanations of another GNN explainers, for example \cite{llmexplainer2024} uses LLMs to evaluate explanations of an explainer and its scores guide the weighted gradients used for optimizing the explainer through Bayesian Variation Inference to address learning bias. Similarly, to improve the understandability of explanations, \cite{tagexplainer2024} first trains a pseudo-label generator LLM that takes explanations from an external GNN explainers and generates pseudo-labels for them. The generator is tuned in an expert iteration procedure with tailored objectives for faithfulness and brevity, and the optimized generator generates pseudo-labels to fine-tune an explainer LLM to build end-to-end models. 

\textbf{\textit{LLM explanation to enhance understanding.}} One advantage of LLMs is that they can generate NLE for GNNs, which are more human-interpretable. \cite{llmgce2024} proposes a method for generating counterfactual explanations, which uses an autoencoder to construct counterfactual graph topologies from LLM-generated counterfactual text pairs. To mitigate hallucinations, it employs a dynamic feedback mechanism that prompts the LLM to refine its initial outputs. Lately, \cite{graphxain2025} proposed prompting an LLM to narrate the explanation subgraphs and the feature importance matrices generated by a GNN explainer, which describe the relationships between neighboring nodes. Our work produces representation-aware LLM explanations without an external explainer. \textit{We present additional related work in Appendix \ref{app:additional_RW}.}

\section{Problem formulation}\label{sec:problem_formulation}

We define a graph as $G = (V, E, X)$, where $V$ is the set of nodes, $E = \{(u, v) \mid u, v \in V\}$ is the set of edges, and $X = \{x_v \mid v \in V\}$ is the node feature matrix, with $x_v \in \mathbb{R}^d$ representing the feature vector of node $v$. We denote by $\mathcal{G}$ the set of graphs in a dataset, and by $\mathcal{V}$  the set of all nodes across the graphs in $\mathcal{G}$.


\begin{defn}[Text-Attributed Graph (TAG)]
A TAG is defined as a graph 
\[
G = (V, E, T)
\]
where $V$ is the set of nodes, $E \subseteq V \times V$ is the set of edges (undirected or directed), $T: V \rightarrow \mathcal{D}$ is a function that assigns a textual document or sequence to each node, where $\mathcal{D}$ is a corpus of natural language texts (e.g., sentences, paragraphs).
\end{defn}
Each node $v \in V$ is thus associated with a text document $T(v)$, and the graph structure $E$ represents semantic, relational, or structural links between nodes. TAGs enable learning from both the textual content and the graph's topology. In this work, we focus on the node classification task.

\textit{Node Classification.} Given a dataset $\mathcal{G}$ and a graph $G = (V, E, T) \in \mathcal{G}$, where every node $v\in V$ has a label $y_v \in \mathcal{Y}$ belonging to one of $C$ classes ($\mathcal{Y} = \{Y_i\}_{i=1}^{C}$), the node classification task consists in training a GNN $\Phi$ such that $\Phi(x_v,G) = y_v \; \; \forall G \in \mathcal{G}, \forall v \in V$, and $x_v=f_e(T(v))$, where $f_e$ is a text encoder. 


Our method aims to explain the node classification predictions of a GNN using local subgraphs, i.e., small subgraphs around the node being classified \cite{gnn_explainer}. 

\begin{defn}
[Local Factual Explainer]\label{def:localexplanation}
Given a GNN $\Phi$ and a graph $G = (V, E, T) \in \mathcal{G}$, a local factual explainer $\Psi$ is a mapping from $(\mathcal{V},\mathcal{G})$ to $\mathcal{S}_\mathcal{V}(\mathcal{G})$ associated to $\mathcal{V}' \subseteq \mathcal{V}$ verifying:
\begin{equation}
\scalebox{0.95}{$
    \Phi(x_v, \Psi(v, G)) = \Phi(x_v,G) \;
    \forall G \in \mathcal{G}, v \in V \cap \mathcal{V}' 
$}
\label{eq:localexplanation}
\end{equation}

where $\mathcal{S}_v(G) \in \mathcal{S}_\mathcal{V}(\mathcal{G})$ denotes the set of all subgraphs of $G$ that contain $v$. $\mathcal{V}'$ corresponds to the set of nodes over all graphs for which the equation is verified, and in this case, the explanation $\Psi(v, G)$ is called faithful.

\end{defn}

Notice that a complete local factual explainer always exists, as $\Psi = \mathrm{Id}$ (identity function) verifies Equation~\ref{eq:localexplanation} for every node. However, such explainers are rarely useful in practice because they do not produce concise, interpretable subgraphs. Therefore, the explanation subgraph size is an important metric, as formally defined in Section~\ref{subsection:metrics}.

Although Definition \ref{def:localexplanation} necessitates faithfulness of the explanation graphs and the existing methods that build explanations for GNNs mostly follow this definition, there are two major problems: 
\begin{itemize}
\item It does not necessarily generate concise explanations. Generally, a good factual explanation subgraph is sparse \cite{surveyonexplainability2023, explainabilitytaxonomysurvey2023}, containing only the most important graph elements: nodes, edges, and node features. 

\item More importantly, this problem definition does not account for human-level interpretability. For example, retrieving an explanation subgraph is common \cite{gnn_explainer, pgexplainer}, but it lacks sufficient information for understanding, which limits its usefulness in practice.

\end{itemize} 
These criteria align with the goals of Explainable AI (XAI). Consequently, we aim to develop a method that enhances interpretability while maintaining conciseness.

\begin{figure*}[h]
    \centering
    \vspace{-2mm}
    \includegraphics[width=0.9\textwidth]{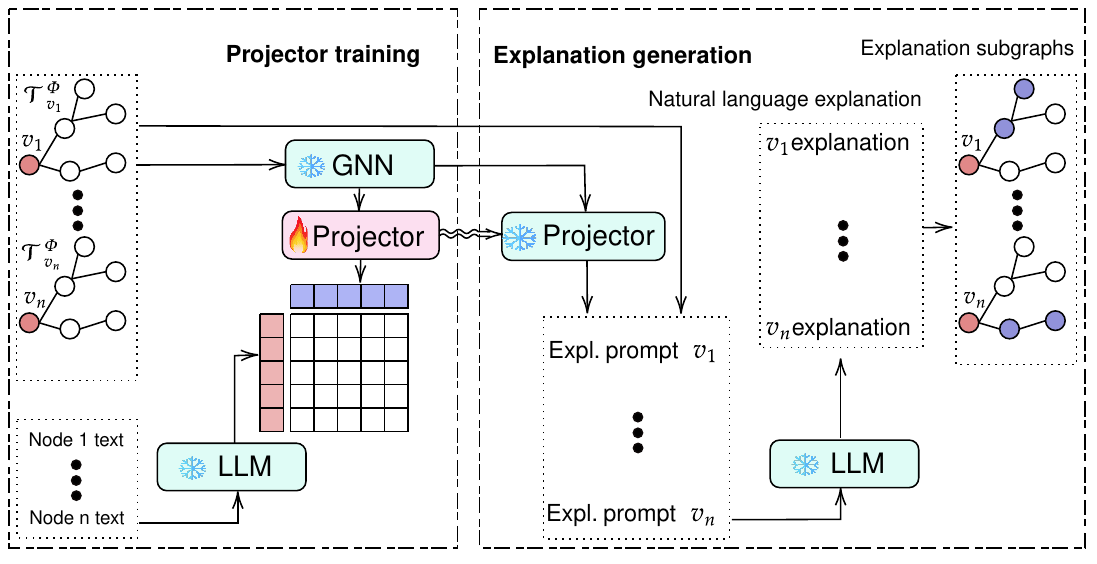} 
    \vspace{-2mm}
    \caption{Illustration of \methodname's framework. First, the projector is trained to align GNN node embeddings with the LLM’s embedding space. Next, hybrid prompts are constructed by interleaving projected embeddings (as soft prompts) with natural language tokens. These prompts are then fed to the LLM to produce natural language explanations, which are converted into explanation subgraphs.}
    \label{fig:method}
    \vspace{-2mm}
\end{figure*}
\section{Our Proposed Method: \methodname}

To explain a trained GNN’s predictions, our method projects node embeddings from the GNN’s latent space into the LLM’s embedding space. By injecting these embeddings as soft prompts interleaved with natural language, we enable the LLM to generate context-aware explanations grounded in the model's internal representations. Our approach consists of three parts: training the projector, constructing the hybrid prompt, and generating the explanation subgraph (Figure \ref{fig:method}).

\subsection{Projection from the GNN}
To help the LLM interpret the GNN, we developed a framework that aligns the LLM with the GNN’s view of the computation tree. Specifically, we use a projector to map GNN node embeddings directly into the LLM's token space.

Given a GNN $\Phi$ trained on $\mathcal{G}$ for node classification on the node set $\mathcal{V}$, for a given graph $G=(V, E, T) \in \mathcal{G}$ and node $v \in V \subseteq \mathcal{V}$, we define $f_\Phi(v)$ to be the GNN embedding of node $v$ in graph $G$, i.e the feature vector of $v$ before the last prediction layer. The projector aligns GNN embeddings with natural language context, effectively bridging two distinct embedding spaces.

Generally speaking, this approach has similarities with other multi-modal alignment frameworks. For instance, in \cite{radford2021learningtransferablevisualmodels}, the authors use contrastive training to align image and text embeddings in a shared space, while in \cite{tsimpoukelli2021multimodalfewshotlearningfrozen}, images are projected directly into a frozen language model’s embedding space to enable coherent language understanding. Similarly, our designed projector learns to map GNN embeddings into soft prompt tokens that serve as input to an LLM, effectively bridging the GNN’s latent space and the LLM’s token-level semantics.

Formally, we define a projector as a function $\Pi: \mathbb{R}^m \to \mathbb{R}^{k \times h}$, where $m$ is the dimension of the GNN embedding $f_\Phi(v) \in \mathbb{R}^m$, $k$ is the number of soft-prompt tokens, and $h$ is the hidden dimension of the LLM's token embeddings. We write $\Pi(f_\Phi(v)) = Z_v \in \mathbb{R}^{k \times h}$. $\Pi$ optimizes two losses:

\begin{enumerate}
    \item \textbf{Context alignment loss:} This encourages the average soft token representation to align with the LLM embedding of the natural language text associated with the node:
    $$\mathcal{L}_{\text{context}} = -\mathbb{E}_{v \in \mathcal{V}} \left[ \text{cos}\left(\bar{Z}_v, \; LLM(v)\right) \right] $$
    where $\bar{Z}_v = \frac{1}{k} \sum_{i=1}^{k} Z_v^{(i)}$ is the mean-pooled projector output, $\text{cos}(.,.)$ denotes cosine similarity, and $LLM(v) \in \mathbb{R}^h$ is the normalized embedding of the text associated with node $v$ obtained from the LLM embedding space.
    
    \item \textbf{Contrastive loss with GNN embeddings:} This encourages soft prompt representations to preserve the similarity structure of the GNN embeddings as follows:
    
   $$\mathcal{L}_{\text{contrast}} = -\frac{1}{|\mathcal{V}|} \sum_{v \in \mathcal{V}} \sum_{u \in \mathcal{V}} p_{vu}^\Phi \log p_{vu}^{\Pi} $$
    where
   $$ p_{vu}^\Phi = \frac{\exp\left( \text{cos}\left(f_\Phi(v), f_\Phi(u)\right) / \tau \right)}{\sum_{w \in \mathcal{V}} \exp\left( \text{cos}\left(f_\Phi(v), f_\Phi(w)\right) / \tau \right)} $$
   
  $$ p_{vu}^{\Pi} = \frac{\exp\left( \text{cos}\left(\bar{Z}_v, \bar{Z}_u\right) \right)}{\sum_{w \in \mathcal{V}} \exp\left( \text{cos}\left(\bar{Z}_v, \bar{Z}_w\right) \right)}$$
    
    and $\tau > 0$ is a temperature hyperparameter.

    The overall training objective becomes:
$\mathcal{L} = \beta \; \mathcal{L}_{\text{context}} + (1-\beta) \mathcal{L}_{\text{contrast}}$
where $1 \geq \beta \geq 0$.
\end{enumerate}

\subsection{Prompt Construction}\label{sec:prompt_construction}

Given a node $u$, the projector outputs a matrix $\Pi(f_\Phi(u)) = Z_u \in \mathbb{R}^{k \times h}$, which we refer to as a soft prompt embedding. This matrix is directly injected into the LLM's embedding space and interleaved with natural language token embeddings.

For a node $v$ that we wish to explain, we consider its computation tree for GNN $\Phi$, denoted by $\mathcal{T}^\phi_v$. $\mathcal{T}^\phi_v$ is a tree with root node $v$ and depth $L$ (the number of layers of $\Phi$) composed of all paths of length $L$ starting at $v$, concatenated to the root $v$. We construct the LLM input as a sequence of embeddings consisting of: (i) Standard token embeddings for the system prompt and user instruction; (ii) The target node's $k$-token embedding $Z_v$---generated after enclosing the node's text descriptor by text markers; (iii) An enumeration of the nodes in $\mathcal{T}^\phi_v$, where each node is enclosed by text markers and its own soft-prompt embeddings; and (iv) Final query instructions. By treating the GNN embeddings as native tokens, the LLM reasons across both GNN representations and textual features (see Figure \ref{fig:prompt-response} and Appendix \ref{appendix:sample}).

\subsection{Explanation Generation}
\label{section:expl_gen}

First, the LLM predicts whether each node in the computation tree supports or does not support the target node's classification. This partitions the nodes of the computation tree into three categories: (i) Supporting nodes,  (ii) Opposing nodes, and (iii) Neutral nodes, which do not appear in either category. Formally, the LLM generates a natural language explanation $E_v$ based on $\mathcal{T}^\Phi_v$ and the classification of $v$. Given the parsing of $E_v$ we apply a label attribution function $ \chi: \mathcal{T}^\phi_v \to \{-1, 0, 1\}$ for each node $u \in \mathcal{T}^\phi_v$ such that:
\begin{itemize}
    \item $\chi(u) = 1$ if $u$ is cited in $E_v$ as supporting the classification of $v$,
    \item $\chi(u) = -1$ if $u$ is cited as not supporting the classification of $v$,
    \item $\chi(u) = 0$ if $u$ is not mentioned.
\end{itemize}

We consider the induced partition of $\mathcal{T}^\phi_v$:

{\small
\[
\begin{aligned}
S^+_v &= \{ u \in \mathcal{T}^\phi_v \mid \chi(u) = 1 \} \\
S^-_v &= \{ u \in \mathcal{T}^\phi_v \mid \chi(u) = -1 \} \\
S^0_v &= \{ u \in \mathcal{T}^\phi_v \mid \chi(u) = 0 \}
\end{aligned}
\]
}

The final explanation subgraph is $S_v=S^+_v$.

\paragraph{Mitigating Hallucination.} LLM-generated explanations may hallucinate, especially when multiple node embeddings are projected within a single prompt. We address this issue through two complementary mechanisms within our framework:
\begin{itemize}
    \item \textbf{Prompt templating with constrained node sets:} To reduce hallucinations, we partition $\mathcal{T}^\phi_v$ into fixed-size batches and sequentially prompt them to the LLM, instructing it to generate explanations that reference only the nodes in each batch.
    \item \textbf{Post hoc filtering:} After the LLM generates the explanations, any hallucinated nodes---the nodes absent in $\mathcal{T}^\phi_v$---are filtered out in a post-processing step. This filtering ensures that the final explanation remains relevant and grounded in the actual GNN model's predictions, as detailed in Section \ref{section:expl_gen}.
\end{itemize}

These strategies work together to minimize hallucination and enhance the accuracy and reliability of the generated explanations.

\subsection{Faithfulness to \gnn Predictions}\label{sec:faithfulness}
Beyond constraining hallucinations, we also aim to ensure that the generated explanations remain faithful to the \gnn's predictions. Accounting for context aligns the GNN embeddings with those of the LLM; however, this alignment alone does not guarantee that the GNN's decisions are faithful to the LLM's. We address faithfulness in two ways.

First, we explicitly include the \gnn's predicted label for the target node in the input prompt (Sec.~\ref{sec:prompt_construction}), so that the LLM conditions its reasoning on the model's actual decision. This provides the LLM with a clear natural-language interpretation of the prediction target.



Second, we introduce \methodnameplus, which directly optimizes soft prompts to ensure faithfulness to the \gnn's predictions. Assume the \gnn is a predictive model $\Phi=h \circ g$ where $g:G \rightarrow \mathcal{E}$ is a message-passing encoder and $h: \mathcal{E} \rightarrow Y$ is a decoder s.t. $\mathcal{E} \in \mathbb{R}^m$ and $Y \in \mathbb{R}^C$ are the latent and output spaces, and $C$ is the number of classes. If $\Phi$ is well optimized, the conditional entropy $H(Y\mid \mathcal{E})$ is low due to the pretraining stage, which is equivalent to increasing the mutual information $I(Y;\mathcal{E})$. Therefore, a good projector $\Pi$ has to not only align the \gnn embeddings with the LLM embeddings, but also have the projections share high mutual information with the output of the \gnn.

Along with the previous alignment objectives $\mathcal{L}_{\text{context}}$ and $\mathcal{L}_{\text{contrast}}$, we further consider increasing the mutual information $I(\hat{y}_v;Z_v)$, where $\hat{y}_v = h(g(v)) \in \mathbb{R}^{C}$ by minimizing the InfoNCE loss \cite{oord2019representationlearningcontrastivepredictive}. Since $Z_v$ and $\hat{y}_v$ do not live in the same space, we use another projector $\Pi':\mathbb{R}^{k \times h} \rightarrow \mathbb{R}^{C}$ to back-propagate through the original projector $\Pi: \mathbb{R}^m \to \mathbb{R}^{k \times h}$ the faithfulness to the prediction. Therefore, writing $Z'_v = \Pi'(Z_v)$, the final objective for \methodnameplus becomes:

\vspace{-0.1cm}

{\small
\begin{align*}
&\mathcal{L} = \beta_1 \; \mathcal{L}_{\text{context}} + \beta_2 \mathcal{L}_{\text{contrast}} + \beta_3\mathcal{L}_{\text{InfoNCE}} \text{ where} \nonumber\\
&\mathcal{L}_{\text{InfoNCE}}= \nonumber\\
&- \mathbb{E} \left[ \log \frac{\exp\left(\frac{\text{cos}(Z'_v, \hat{y}_v)}{\tau}\right)}{\exp\left(\frac{\text{cos}(Z'_v , \hat{y}_v)}{\tau} \right) + \underset{u\ \neq v}\sum_{}^{} \exp\left(\frac{\text{cos}(Z'_v, \hat{y}_u)}{\tau}\right)} \right] \nonumber\\
\end{align*}}

\begin{table*}[t]
    \centering
    \footnotesize
    
    \resizebox{\textwidth}{!}{
        \begin{tabular}{lcccccccc}
            \toprule
            & \multicolumn{2}{c}{\textsc{Amazon}} 
            & \multicolumn{2}{c}{\textsc{Cora}} 
            & \multicolumn{2}{c}{\textsc{Liar}} 
            & \multicolumn{2}{c}{\textsc{WikiCS}} \\
            
            \cmidrule(lr){2-3} \cmidrule(lr){4-5} \cmidrule(lr){6-7} \cmidrule(lr){8-9}
            
            & Fidelity (\%) & Size & Fidelity (\%) & Size & Fidelity (\%) & Size & Fidelity (\%) & Size \\
            \midrule
            
            \textsc{GNNExplainer} & 
            94.5 $\pm$ 1.5 & 5.71 $\pm$ 0.01 & 
            96.5 $\pm$ 0.6 & 17.91 $\pm$ 0.01 & 
            100.0 $\pm$ 0.00 & 474.01 $\pm$ 0.00 & 
            99.0 $\pm$ 0.0 & 1930.89 $\pm$ 0.0 \\
            
            \textsc{PGExplainer} & 
            73.8 $\pm$ 5.6 & 4.52 $\pm$ 0.13 & 
            82.0 $\pm$ 1.0 & 8.96 $\pm$ 0.40 & 
            100.0 $\pm$ 0.0 & *1.0 & 
            90.6 $\pm$ 0.0 & 295.89 $\pm$ 0.0 \\
            
            \textsc{TAGExplainer} & 
            65.1 $\pm$ 4.4 & 1.03 $\pm$ 0.00 & 
            78.5 $\pm$ 0.9 & 1.06 $\pm$ 0.00 & 
            100.0 $\pm$ 0.0 & 1.31 $\pm$ 0.00 & 
            76.4 $\pm$ 1.1 & 1.51 $\pm$ 0.00 \\
            
            \textsc{Node} & 
            67.8 $\pm$ 2.7 & 1.00 $\pm$ 0.00 & 
            77.5 $\pm$ 2.6 & 1.00$\pm$ 0.00 & 
            100.0 $\pm$ 0.0 & 1.00 $\pm$ 0.00 & 
            72.3 $\pm$ 0.0 & 1.00 $\pm$ 0.0 \\
            
            \textsc{Random} & 
            91.0 $\pm$ 1.1 & 7.28 $\pm$ 0.02 & 
            93.7 $\pm$ 0.5 & 18.67 $\pm$ 0.01 & 
            100.0 $\pm$ 0.0 & 469.36 $\pm$ 0.03 & 
            93.3 $\pm$ 0.0 & 994.76 $\pm$ 0.0 \\
            
            \textsc{LLM} & 
            86.0 $\pm$ 1.2 & 4.90 $\pm$ 0.23 & 
            94.5 $\pm$ 1.0 & 3.27 $\pm$ 0.31 & 
            100.0 $\pm$ 0.0 & 2.22 $\pm$ 0.04 & 
            91.4 $\pm$ 1.0 & 21.19 $\pm$ 0.30 \\
            
            \midrule
          \methodnameplus & 
            92.0 $\pm$ 1.8 & 4.24 $\pm$ 0.50 & 
            85.8 $\pm$ 1.8 & 2.10 $\pm$ 0.11 & 
            100.0 $\pm$ 0.0 & 1.10 $\pm$ 0.08 & 
            86.3 $\pm$ 1.7 & 12.44 $\pm$ 0.27 \\
            
            \methodname & 
            91.1 $\pm$ 2.3 & 4.25 $\pm$ 0.54 & 
            86.5 $\pm$ 0.0 & 2.07 $\pm$ 0.07 & 
            100.0 $\pm$ 0.0 & 1.11 $\pm$ 0.08 & 
            86.5 $\pm$ 0.0 & 12.80 $\pm$ 0.0 \\
            
            \bottomrule
        \end{tabular}
    }
    \caption{Performance comparison with baseline models on the node classification task. Higher fidelity and lower size are better. *The explainer returns all-zero masks on \liar, we consider its output to be the central node only.}
    \label{tab:maintable}
\end{table*}

\begin{figure}[ht]
\centering
\scriptsize
\setlength{\tabcolsep}{3pt}


\vspace{-1pt}
\begin{minipage}[t]{0.48\linewidth}
\vspace{0pt}
\begin{tcolorbox}[tightbox,tighttitle,title={LLM Prompt}]
\RaggedRight
You are analyzing Amazon product reviews and their co-purchase neighborhoods to understand why it has been classified under a specific category. Your task is to determine which of the neighboring products does (not) support the classification of the target product. 

\textbf{Target Product ID:} 160\\
\textbf{Predicted Category:} \PredCat{Sports \& Outdoors}\\[3pt]

\textbf{Product Embedding Representation:}\\
\BeginTK\ \EmbedText{PROJECTED EMBEDDING}\ \EndTK\\[3pt]

\textbf{Neighboring Products:}\\Each Product below is described by keywords.\\[3pt]
\textbf{Product 3:} \NeighborBeginTK\ \EmbedText{PROJECTED EMBEDDING}\ \NeighborEndTK\\
\textbf{Product 4:} \NeighborBeginTK\ \EmbedText{PROJECTED EMBEDDING}\ \NeighborEndTK\\
\textbf{Product 5:} \NeighborBeginTK\ \EmbedText{PROJECTED EMBEDDING}\ \NeighborEndTK\\
\textbf{Product 10:} \NeighborBeginTK\ \EmbedText{PROJECTED EMBEDDING}\ \NeighborEndTK\\
\textbf{Product 12:} \NeighborBeginTK\ \EmbedText{PROJECTED EMBEDDING}\ \NeighborEndTK\\

\end{tcolorbox}
\end{minipage}
\hfill
\begin{minipage}[t]{0.48\linewidth}
\vspace{0pt}
\begin{tcolorbox}[tightbox,tighttitle,title={LLM Response}]
\RaggedRight
\begin{tabular}{@{}p{0.10\linewidth}p{0.68\linewidth}p{0.15\linewidth}@{}}

\textbf{P3} & Summary: This product seems to be a fitness accessory, possibly a yoga mat or a resistance band, given its mention of ``workout'' and ``stretching''. & \SupportYes \\
\textbf{P4} & Summary: This product appears to be a nutrition supplement, possibly a protein powder or a vitamin, given its mention of ``muscle'' and ``energy''. & \SupportNo \\
\textbf{P5} & Summary: This product seems to be a fitness tracker or a smartwatch, given its mention of ``fitness'' and ``activity tracking''. & \SupportYes \\
\textbf{P10} & Summary: This product seems to be a health and wellness product, possibly a meditation app or a mindfulness journal, given its mention of ``relax'' and ``meditate''. & \SupportNo \\
\textbf{P12} & Summary: This product appears to be a sports accessory, possibly a water bottle or a towel, given its mention of ``hydrate'' and ``dry off''. & \SupportYes \\
\end{tabular}
\end{tcolorbox}
\end{minipage}

\vspace{4pt}
\begin{tcolorbox}[tightbox,tighttitle,title={Instructions}]
\RaggedRight
For each neighboring product: \\
1. Summarize the similarity between supporting products in \emph{one sentence}. \\
2. Clearly state whether this product supports the classification of the Target Product into category \PredCat{Sports \& Outdoors}. \\[2pt]
\textbf{Format:}\\
\texttt{Product <ID>:}\\
\texttt{Summary: <One sentence>.}\\
\texttt{Support: YES or NO — Does this product support classification into \PredCat{Sports \& Outdoors}?}\\[2pt]
Base reasoning only on the keywords and proximity to the target product.
\end{tcolorbox}

\vspace{-3pt}
\caption{Left: prompt with category and embeddings highlighted. Right: full model response with summaries in the middle and YES/NO verdicts aligned on the right.} 
\label{fig:prompt-response}
\end{figure}

\vspace{-6mm}
\section{Experiments}
\label{section:experiments}

We demonstrate the effectiveness of \methodname in various settings. Our code is publicly available at: \url{https://github.com/pbaghershahi/GSPELL.git}.

\subsection{Experimental setup}
\label{subsection:metrics}

\textbf{Evaluation Metrics. }Having an explainer $\Psi$ of GNN $\Phi$, a dataset $\mathcal{G}$ of $\mathcal{V}$ nodes, and denoting explanation subgraph by $S_v=\Psi(v,G)$, we use the following metrics in the experiments:

\noindent
\paragraph{1) Fidelity:}
Fidelity captures how well an explainable model reproduces the GNN model's logic in its predictions. Mathematically, we have: 
\vspace{-0.1cm}
$$Fid = \frac{1}{|\mathcal{V}|} \sum_{G \in \mathcal{G}, v \in V} \mathbb{I}[\Phi(v, S_v) = \Phi(v,G)]$$
\vspace{-0.1cm}
where {\small $\mathbb{I}[\cdot]$} is the indicator function.

\noindent
\paragraph{2) Size:} Smaller explanation sizes are easier to interpret. The explanation size measures the compactness of the explanation subgraph, i.e., the number of nodes in the explanation, as: 
\vspace{-0.1cm}
$$Size = \frac{1}{|\mathcal{V}|} \sum_{G \in \mathcal{G}, v \in \mathcal{V}} |S_v|$$
\vspace{-0.1cm}

For a more comprehensive evaluation, we adopt the perturbation-based framework of recent natural language explanation literature \cite{fathfullperturbation12023, fathfullperturbation12024}.

\noindent
\paragraph{3) Comprehensiveness: }
Comprehensiveness measures the drop in prediction accuracy after removing the explanation subgraph. Let $G \setminus S_v$ denote the graph obtained by removing the nodes in $S_v$. Then, we define comprehensiveness as: 
\begin{align}
Comp. =
\nonumber \frac{1}{|\mathcal{V}|}
\sum_{G \in \mathcal{G},\, v \in V}
\mathbb{I}\!\left[\Phi(x_v,G)=y_v\right]
-\\
\nonumber \mathbb{I}\!\left[\Phi\!\left(x_v,G \setminus S_v\right)=y_v\right],
\end{align}
\vspace{-0.6cm}

\noindent
\paragraph{4) Alignment: }
Alignment evaluates whether the nodes identified as opposing the prediction, the set $S_v^{-}$, are indeed unsupportive. We define alignment as the drop in prediction accuracy after removing these nodes: 
\vspace{-0.1cm}
\begin{align}
Align. =
\nonumber \frac{1}{|\mathcal{V}|}
\sum_{G \in \mathcal{G},\, v \in V}
\mathbb{I}\!\left[\Phi(v,G)=y_v\right]
-\\
\nonumber \mathbb{I}\!\left[\Phi\!\left(v,G \setminus S_v^{-}\right)=y_v\right].
\end{align}

\vspace{-0.3cm}
\noindent
\paragraph{5) Random Baseline: }
The random baseline measures the drop in prediction accuracy after removing a random node set, the same size as the explanation subgraph. Let $S_v^{\mathrm{rand}} \subseteq T_v^\phi$ be a uniformly sampled subset such that $|S_v^{\mathrm{rand}}| = |S_v|$, we define: 

\vspace{-0.3cm}
\begin{align}
Rand. =
\nonumber \frac{1}{|\mathcal{V}|}
\sum_{G \in \mathcal{G},\, v \in V}
\mathbb{I}\!\left[\Phi(v,G)=y_v\right]
-\\
\nonumber \mathbb{I}\!\left[\Phi\!\left(v,G \setminus S_v^{\mathrm{rand}}\right)=y_v\right].
\end{align}

Overall, higher values are preferred for Fidelity and Comprehensiveness, but lower values are preferred for Size and Alignment. The Random Baseline serves as a reference, where we expect Comprehensiveness to be significantly higher than the Random Baseline.

\noindent
\textbf{Datasets.} We use real-world TAG datasets to evaluate the performance of our method: \cora~\cite{cora_dataset_src}, \wiki~\cite{mernyei2022_wikics}, \amazonprod~\cite{feng2024_taglas}, and  \liar~\cite{wang2017liarliarpantsfire}. Additional details of the datasets are provided in Appendix \ref{app:dataset-details}.

\noindent
\textbf{Models.} We consider a GCN~\cite{kipf2017semisupervised} as the base GNN model for the node classification task, and the LLM model is Meta's Llama-3.1-8B-Instruct, which remains frozen in our experiments. Full experimental details are provided in Appendix \ref{app:add-implem-details}. We provide experiments with other LLM models in Appendix \ref{app:LLM_Backbone} and for different GNN architectures in Appendix \ref{app:GNN_Backbone}.

\begin{table*}[t]
\centering
\renewcommand{\arraystretch}{1}
\small
\begin{tabular}{lcccccc}
\toprule
\multirow{2}{1.2in}{Dimension} &
\multicolumn{3}{c}{\cora} &
\multicolumn{3}{c}{\amazonprod} \\
\cmidrule(lr){2-4} \cmidrule(lr){5-7}
 & $M_1$$\uparrow$ & $M_2$$\uparrow$ & $\Delta$ 
 & $M_1$$\uparrow$ & $M_2$$\uparrow$ & $\Delta$ \\
\midrule

Understandability & $2.9_{\scriptstyle \pm 1.23}$ & $3.2_{\scriptstyle \pm 1.45}$ & $0.3$ 
& $3.33_{\scriptstyle \pm 1.46}$ & $3.25_{\scriptstyle \pm 1.33}$ & $-0.08$ \\
Trustworthiness   & $2.7_{\scriptstyle \pm 1.10}$ & $3.1_{\scriptstyle \pm 1.06}$ & $0.4$ 
& $3.30_{\scriptstyle \pm 1.56}$ & $3.05_{\scriptstyle \pm 1.41}$ & $-0.25$ \\
Insightfulness    & $3.3_{\scriptstyle \pm 0.74}$ & $3.1_{\scriptstyle \pm 0.95}$ & $-0.2$ 
& $2.95_{\scriptstyle \pm 1.34}$ & $3.25_{\scriptstyle \pm 1.17}$ & $0.30$ \\
Satisfaction      & $2.2_{\scriptstyle \pm 1.16}$ & $2.7_{\scriptstyle \pm 0.92}$ & $0.5$ 
& $3.08_{\scriptstyle \pm 1.46}$ & $3.05_{\scriptstyle \pm 1.36}$ & $-0.03$ \\
Confidence        & $3.0_{\scriptstyle \pm 1.49}$ & $3.3_{\scriptstyle \pm 1.49}$ & $0.3$ 
& $3.13_{\scriptstyle \pm 1.47}$ & $3.03_{\scriptstyle \pm 1.33}$ & $-0.10$ \\
Convincingness    & $3.3_{\scriptstyle \pm 1.57}$ & $3.7_{\scriptstyle \pm 1.64}$ & $0.4$ 
& $3.08_{\scriptstyle \pm 1.46}$ & $2.93_{\scriptstyle \pm 1.49}$ & $-0.15$ \\
Communicability   & $3.0_{\scriptstyle \pm 1.60}$ & $3.1_{\scriptstyle \pm 1.41}$ & $0.1$ 
& $2.88_{\scriptstyle \pm 1.28}$ & $2.93_{\scriptstyle \pm 1.23}$ & $0.05$ \\
Usability         & $3.3_{\scriptstyle \pm 1.43}$ & $3.4_{\scriptstyle \pm 1.42}$ & $0.1$ 
& $2.75_{\scriptstyle \pm 1.39}$ & $2.80_{\scriptstyle \pm 1.22}$ & $0.05$ \\

\bottomrule
\end{tabular}
\caption{Expert qualitative evaluation of the explanations from \methodname and GNNExplainer on \cora and \amazonprod. $M_1$ denotes the mean scores for GNNExplainer and $M_2$ the mean scores for \methodname.}
\label{tab:qualitative_results}
\end{table*}

\begin{table*}[t]
    \centering
    \footnotesize
    
    \resizebox{\textwidth}{!}{
        \begin{tabular}{lccccccccccccccc}
            \toprule
            & \multicolumn{5}{c}{\textsc{Amazon}} 
            & \multicolumn{5}{c}{\textsc{Cora}} 
            & \multicolumn{5}{c}{\textsc{WikiCS}} \\
            
            \cmidrule(lr){2-6} \cmidrule(lr){7-11} \cmidrule(lr){12-16}
            
            & F (\%) $\uparrow$ & C (\%) $\uparrow$ & R (\%) & A (\%) $\downarrow$ & S $\downarrow$ 
            & F (\%) $\uparrow$ & C (\%) $\uparrow$ & R (\%) & A (\%) $\downarrow$ & S $\downarrow$ 
            & F (\%) $\uparrow$ & C (\%) $\uparrow$ & R (\%) & A (\%) $\downarrow$ & S $\downarrow$  \\
            \midrule
            
            \textsc{GNNExplainer} & 
            94.5 & 14.7 & 6.9 & - & 5.71 &   
            96.5 & 15.8 & 7.8 & - & 17.91 &  
            99.0 & 13.1 & 13.1 & - & 1930.89 \\         

            
            \methodname & 
            91.1 & 8.2 & 4.8 & 3.0 & 4.25 &   
            86.5 & 5.8 & 1.2 & 7.8 & 2.07 &   
            86.5 & 4.20 & 0.0 & 4.7 & 12.80 \\  
            
            \methodnameplus & 
            92.0 & 8.3 & 3.5 & 5.0 & 4.24 &   
            85.8 & 6.2 & 1.5 & 8.2 & 2.10 &   
            86.3 & 3.8 & 0.2 & 5.8 & 12.44 \\   
            
            \bottomrule
            \vspace{-5mm}
        \end{tabular}
    }
    \caption{Perturbation-based evaluation of faithfulness. \methodname shows a superior trade-off between faithfulness and conciseness, achieved by isolating critical nodes for human-centered explanations in natural language.}
    \label{tab:faithfulness_eval}
\end{table*}

\noindent
\textbf{Baselines.}
We consider the following \textit{five} baselines: \textbf{(1) Node:} This baseline uses only the target node itself as the explanation subgraph. This represents the minimal possible context and serves as a lower bound on explanation size.
\textbf{(2) Random}: We introduce a randomized baseline that selects a subgraph around each target node as its explanation subgraph. Specifically, for each node, we sample a subgraph of half the size of its computation tree. The results are averaged over 5 samples. \textbf{(3-5) Graph Explainers}: We compare \methodname with GNNExplainer~\cite{gnn_explainer}, PGExplainer \cite{pgexplainer} as subgraph-based explainers, which do not involve any LLM-based component. We also include the recent LLM-based graph explanation method, TAGExplainer~\cite{tagexplainer2024}.

\subsection{Quantitative Results}

Table \ref{tab:maintable} shows that the Node baseline is relatively strong, highlighting the critical role of textual attributes in GNN predictions. While GNNExplainer often reaches the highest fidelity, this metric alone can be misleading. Since GNNs aggregate neighborhood data, high fidelity can be trivially achieved by simply outputting the entire induced subgraph.

In homophilic graphs, where neighbors often share the same category, selecting a broad neighborhood effectively replicates the GNN’s original context. This is evidenced by the Random baseline, which achieves fidelity scores comparable to GNNExplainer. In these settings, attaining high fidelity is trivial if the explanation subgraph is large.

Across all datasets, \methodname achieves a superior fidelity-size trade-off: despite marginally lower fidelity than GNNExplainer and the LLM baseline, it produces significantly smaller subgraphs. Unlike Random or Node baselines, which either inflate size or restrict context, \methodname leverages projected GNN embeddings to distill rich textual semantics into concise, interpretable explanations.

Comparing \methodnameplus to \methodname highlights the impact of the InfoNCE objective: while it further reduces explanation size, it occasionally lowers fidelity. The objective's effectiveness scales with the number of negative samples; for instance, it outperforms on \amazonprod compared to \cora because a higher class count provides a tighter mutual information bound. 

\paragraph{Perturbation-based Evaluation.} To further assess the faithfulness of the generated explanations, we adopt perturbation-based evaluation. 

Table \ref{tab:faithfulness_eval} demonstrates that \methodname optimizes the trade-off between conciseness and faithfulness. While GNNExplainer shows high fidelity, it relies on brute-force coverage—often selecting nearly entire neighborhoods. For instance, on Cora, GNNExplainer averages 18 nodes per explanation, whereas \methodname averages 2. This distinction is further highlighted on WikiCS: GNNExplainer produces excessively large, unreadable subgraphs, whereas \methodname identifies high-impact nodes within significantly smaller, interpretable structures. Moreover, \methodname achieves a comprehensiveness of 2–5$\times$ greater than the random drop baseline, outperforming GNNExplainer’s 2–3$\times$ margin. This demonstrates that \methodname maintains a superior signal-to-noise ratio, making human-readable narratives.

Unlike baselines that provide binary-mask explanations, \methodname’s 3-way classification—Support, Oppose, and Neutral—uniquely enables the Alignment metric to differentiate between contradicting evidence and irrelevant noise to detect hallucinations. Across all datasets, \methodname's Alignment scores are low, particularly on Amazon, confirming successful detection of irrelevant nodes. 

\subsection{Qualitative Evaluation}
\paragraph{Expert Assessment.} We asked experts to evaluate the explanations generated by GNNExplainer and \methodname on the \amazonprod and \cora datasets over 10 node classification tasks. We follow the qualitative evaluation framework provided in \cite{graphxain2025}, and refer to Appendix \ref{app:qualitative_metric} for a full description of the evaluation dimensions. The evaluation framework and setup are presented in Appendix \ref{appendix:human_eval}. 


\begin{table*}[t]
    \centering
    \footnotesize
    
    \resizebox{0.9\textwidth}{!}{
        \begin{tabular}{llcccccccccc}
            \toprule
            & & \multicolumn{2}{c}{$k=1$} 
            & \multicolumn{2}{c}{$k=5$} 
            & \multicolumn{2}{c}{$k=10$} 
            & \multicolumn{2}{c}{$k=20$} 
            & \multicolumn{2}{c}{$k=50$} \\
            \cmidrule(lr){3-4} \cmidrule(lr){5-6} \cmidrule(lr){7-8} \cmidrule(lr){9-10} \cmidrule(lr){11-12}
            
            & & Fidelity (\%) & Size & Fidelity (\%) & Size & Fidelity (\%) & Size & Fidelity (\%) & Size & Fidelity (\%) & Size \\
            \midrule
            
            \multirow{2}{*}{Amazon} & \methodname & 
            92.0 & 4.22 & 91.1 & 4.25 & 88.0 & 1.71 & 85.0 & 1.17 & 82.5 & 1.00 \\
            
            & \methodnameplus & 
            90.5 & 4.45 & 92.0 & 4.24 & 86.0 & 1.45 & 85.0 & 1.13 & 81.0 & 1.01 \\
            \midrule
            
            \multirow{2}{*}{Liar} & \methodname & 
            100.0 & 1.13 & 100.0 & 1.11 & 100.0 & 1.11 & 100.0 & 1.00 & OOM & OOM \\
            
            & \methodnameplus & 
            100.0 & 1.14 & 100.0 & 1.10 & 98.4 & 1.10 & 100.0 & 1.00 & OOM & OOM \\
            \midrule
            
            \multirow{2}{*}{Cora} & \methodname & 
            86.2 & 1.98 & 86.5 & 2.07 & 87.8 & 2.04 & 85.4 & 1.81 & 84.4 & 1.33 \\
            
            & \methodnameplus & 
            85.2 & 1.87 & 85.8 & 2.10 & 88.0 & 2.17 & 87.4 & 2.17 & 84.8 & 1.30 \\
            
            \bottomrule
        \end{tabular}
    }
    \caption{Effect of the number of tokens ($k$) on the performance of \methodname and \methodnameplus. }
    \label{tab:ablation_k}
\end{table*}


As Table \ref{tab:qualitative_results} shows, \methodname's explanations on \cora were more trustworthy and convincing, likely due to the LLM component’s semantic understanding. However, these explanations were seen as less insightful, suggesting more limited exploration beyond the article’s main subject. On \amazonprod, \methodname provided more insightful explanations than GNNExplainer by capturing richer semantic links across product categories, but it was rated lower in trustworthiness and convincingness, indicating less consistency and reliability.

Overall, our method performs better on \cora, where its explanations were viewed as more trustworthy, convincing, and communicable. On the more challenging \amazonprod dataset, GNNExplainer retained an edge in clarity and user confidence, suggesting that dataset complexity and semantic richness shape the relative strengths of LLM-based explanations.

\paragraph{Real Example.} Figure \ref{fig:prompt-response} illustrates a prompting example of \methodname on the Amazon dataset to emphasize the narrative quality and interpretability of the generated explanation. In this instance, the target node corresponds to a product (ID 160), predicted to belong to the ``Sports \& Outdoors'' category. \methodname leverages latent GNN representations via soft prompts to benefit from the LLM's language-generation capabilities while accounting for the local neighborhood of the target node.

\methodname generates node-level rationales from the target node’s computation tree, providing for each neighbor a binary support judgment (YES/NO) indicating whether it supports the target’s classification. Also, it provides a general summary that highlights common patterns across the target’s context, helping to explain the model’s prediction.

\section{Ablation Study and Runtime Analysis}
\label{appendix:additional_exp}

\subsection{Effect of the projection tokens}

Without accounting for GNN internals (embeddings), asking an LLM to generate explanations only based on the input and the predictions of the GNN is theoretically wrong. The main reason is that one can directly substitute the predictive GNN with any arbitrary function that makes predictions based on different reasoning (logic) than the GNN, while still making the same predictions. Therefore, we use the soft prompt generated by the projector for the faithfulness of the explanations.

Soft prompts channel information from the GNN to the LLM, with the token count, $k$, being a critical factor here. Insufficient tokens constrain the information channel, causing the LLM to ignore GNN-derived context in favor of textual priors, thereby increasing hallucination rates. Conversely, a large number of tokens can disrupt coherence during generation. If the soft prompt distribution aligns poorly with the LLM’s embedding space---often due to a suboptimal projection function--increasing $k$ injects out-of-distribution noise into the model. This results in degraded generation quality and increases computational overhead.

We perform an ablation study over different token counts ($k$), shown in Table \ref{tab:ablation_k}. We observe that $k$ acts as a sparsity regulator. On the complex Amazon dataset, low $k$ yields high fidelity with larger explanations, while high $k$ significantly reduces size at the cost of fidelity. This supports our hypothesis that excessive soft tokens inject noise into the LLM. We find an optimal balance at $k=5$, achieving high fidelity with reduced explanation size. In contrast, Liar, a more heterogeneous dataset, benefits from a higher $k$ as it likely requires more tokens to encode complex relationships than standard homogeneous graphs. Conversely, Cora exhibits high robustness in both fidelity and sparsity across all $k$ values, likely due to its higher homophily.

\subsection{Runtime}

We provide the details of running times in Table \ref{tab:runtime}, comparing our method with all baselines. Although \methodname is more time-consuming than other baselines, we note that this is justified by the fact that \methodname, unlike the fast baselines, e.g., PGExplainer and TAGExplainer, generates descriptive, informative explanations in natural language that are understandable by humans.

\begin{table}[t]
    \centering
    \footnotesize
    
    \resizebox{\columnwidth}{!}{
        \begin{tabular}{lcccc}
            \toprule
            & \textsc{Amazon} & \textsc{Cora} & \textsc{Liar} & \textsc{WikiCS} \\
            \midrule
            \textsc{GNNExplainer} & 1.45 & 1.69 & 1.79 & 3.76 \\
            \textsc{PGExplainer} & 0.02 & 0.03 & 0.79 & 0.80 \\
            \textsc{TAGExplainer} & 0.05 & 0.13 & 0.52 & 1.57 \\
            \textsc{LLM} & 33.74 & 18.21 & 12.80 & 61.67 \\
            \textbf{\methodname (Ours)} & 39.65 & 30.77 & 42.41 & 76.76 \\
            \bottomrule
            \vspace{-5mm}
        \end{tabular}
    }
    \caption{Inference runtime per node for explanation generation measured in seconds.}
    \label{tab:runtime}
\end{table}

\section{Conclusions}
In this work, we propose \methodname, a lightweight, post-hoc explanation framework that uses LLMs to generate faithful and human-interpretable explanations for GNN predictions on text-attributed graphs. By projecting GNN embeddings into the LLM space, \methodname enables context-aware, natural language rationales without relying on an external GNN explainer. Our approach achieves a superior trade-off between conciseness and fidelity, as well as greater semantic clarity, compared to prior methods. \methodname demonstrates that LLMs can serve as effective interpreters for GNNs and offers a scalable and intuitive path toward explainability in graph-based applications.


\textbf{Future Work. }Although \methodnameplus incorporates an InfoNCE loss to enhance faithfulness, it does not consistently outperform \methodname. Moreover, the other projector objectives focus on aligning GNN and LLM embedding spaces rather than directly optimizing explanation quality. Designing more principled, explanation-centered objectives is an important direction for future work.

Additionally, as the computation tree grows, the LLM struggles to capture the full graph context. While batch-wise prompting mitigates this issue, it limits global reasoning. Developing prompt compression techniques that preserve structural information could improve the quality of the generated explanations.

\section*{Acknowledgment}
 The authors acknowledge the National Artificial Intelligence Research Resource (NAIRR) Pilot and the Texas Advanced Computing Center (TACC) Vista for contributing to this research.
\clearpage
\section{Ethical considerations}
In this work, we have built methods for generating explanations for graph neural network predictions. We do not foresee any ethical issues from our study. \\

\textbf{Potential Risks.} While \methodname aims to improve the interpretability of GNN predictions, the generated explanations rely on LLM outputs and may not always faithfully reflect the underlying model's true decision process. This could lead to over-trust in the explanations. Also, the use of LLMs introduces the risk of hallucinated or misleading reasoning, despite our mitigation strategies. We encourage careful human oversight when deploying such explanation systems in practice.

\section{Limitations}

We present some limitations of our work:

\begin{itemize}
    \item Our method combines projected GNN embeddings with text in hybrid prompts. When the computation tree is large, processing the full context can lead to hallucinations, as the LLM struggles to distinguish among multiple soft prompt representations, while batch-wise prompting limits its ability to capture a global view.
    \item The explanations generated by \methodname are produced by an LLM conditioned on projected embeddings rather than explicitly tracing the GNN’s computation. Consequently, they may not reflect the true causal factors underlying the model’s prediction, so we do not provide formal guarantees of causal faithfulness.
    \item Although the projector aligns GNN embeddings with the LLM embedding space, its training objectives are not directly tailored for explanation. While we explore an InfoNCE-based objective to improve faithfulness, it does not consistently yield gains, suggesting that more effective explanation-specific objectives are needed.
\end{itemize}

\clearpage
\bibliography{custom}

\clearpage
\appendix

\appendix
\section{Additional Related Work}
\label{app:additional_RW}
Here we provide additional related works that are on the integrated methods of LLMs and GNNs. The integration of LLMs with GNNs aims to leverage the advanced reasoning abilities of LLMs and the extensive background knowledge they acquire, alongside the GNNs' proven capacity to enhance the exploitation of graph structures via message-passing mechanisms \cite{gcn2017, graphsage2017}. The approaches for integrating LLMs and GNNs primarily vary in the sequence of component application \cite{surveyllmgnnkdd2024, explorellmgnn2024}. 

\textbf{\textit{LLMs as prefix and postfix. }}First, there are methods where LLMs are used in the prefix modules for task and feature representation \cite{oneforall2024, gaugllm2024, tape2024, glfusion2025}. This architecture provides greater flexibility, as various tasks can be specified in natural language with text-based graph representations, and LLM modules can remain fixed. On the other hand, the second class of methods uses GNNs as the prefix \cite{canlarge2023, llaga2024, askgnn2024, graphgpt2024}, which is better suited for capturing the graph topology and its intricate relationships and uses LLMs for prediction. This approach also has some flexibility, as the LLM can output embeddings, prediction scores, or textual sequences to address various downstream problems specifically for generation tasks \cite{canlarge2023, llaga2024, askgnn2024, graphgpt2024}. 

\textbf{\textit{Independent LLMs and joint LLM-GNNs.}} Additionally, there is a body of work that employs LLMs independently to tackle graph-related tasks \cite{structurellmgnn2024, talklikegraph2024, zerog2024}. Approaches following this paradigm benefit from LLMs' generalization capabilities at both the input and output levels and can potentially avoid training learnable parameters by fixing the LLM component and relying on input prompts. Finally, there is a group of works which employ \gnns and LLMs jointly through soft prompting, which necessitates alignments between their embeddings while enhancing LLMs reasoning by the \gnn captured graph structural features \cite{canwesoftprompt2024, llmaszeroshot2024, graphneuralprompting2023}. However, these works do not address the GNN explanation problem.

\textbf{Explainers for GNNs.} Several algorithms have been proposed in the literature to enhance the interpretability of GNNs. The majority of existing explainers concentrate on instance-level explanations. Instance-level (or local) explainers operate on individual graphs, identifying influential components—such as subgraphs—that most strongly impact the model’s prediction~\cite{gnnexplainer,pgexplainer,rgexplainer,graphlime,yuan2022explainability,Cfgnnexplainer,cff, gem, rcexplainer,braincf, chhablani2024game,moleculecf,induce,goat}. However, this localized perspective constrains their ability to uncover global patterns leveraged by GNNs across multiple graphs, as well as how these patterns integrate into a unified decision-making process. Research on global GNN explainers remains relatively limited~\cite{xgnn,gcfexplainer,xuanyuan2023global,glgexplainer,armgaan2024graphtrail,fournier2025comrecgc}. For instance, XGNN~\cite{xgnn} and GNNInterpreter~\cite{gnninterpreter} adopt generative modeling approaches, producing graphs that most strongly correspond to a given class label. For more details, please refer to this survey on GNN explainers \cite{surveyonexplainability2023}.




\begin{table}[h]

    \centering
\resizebox{0.45\textwidth}{!}{
\begin{tabular}{c|ccccc}
\toprule
         Dataset   & \# Nodes & \# Edges & \# Categories & \# Features \\ \hline
\cora & 2,708 & 5,429 & 7 & 2000 \\
\wiki & 11701 & 148555 & 10 & 300  \\
\amazonprodsub & 1000 & 1397 & 47 & 2000  \\
\liar & 13293 & 32443 & 7 & 387  \\
\bottomrule
\end{tabular}
}
\caption{Statistics of the graph datasets.}    
\label{tab:dataset_stats}
\end{table}

\section{Datasets}
\label{app:dataset-details}

\begin{table*}[ht]
    \centering
    \resizebox{\textwidth}{!}{
    \begin{tabular}{|l|l|l|}
    \hline
    \textbf{Dataset} & \textbf{Description \& Relevance} & \textbf{Why It Fits Our Work} \\
    \hline
    \textbf{CORA} & Citation network of CS papers with bag-of-words features. & Standard benchmark for node classification on text-rich graphs. \\
    \hline
    \textbf{WIKICS} & Wikipedia CS articles linked by hyperlinks, with GloVe embeddings. & Large, semantic-rich graph ideal for explanation evaluation. \\
    \hline
    \textbf{LIAR} & Fake news detection graph combining statements, speakers, topics. & Challenging, heterogeneous graph testing method adaptability. \\
    \hline
    \textbf{AMAZON} & Product co-purchase network with 47 categories. & Real-world, diverse e-commerce graph to test scalability. \\
    \hline
    \end{tabular}
    }
    \caption{Dataset Descriptions and Relevance to Our Work}
    \label{tab:datasets}
\end{table*}

\subsection{Datasets Details}

We describe in detail the datasets that we used in our evaluation below. The basic statistics about these datasets are provided in Table \ref{tab:dataset_stats}.
\begin{enumerate}
    \item \cora : It is a citation network, in which nodes represent computer science research papers, and each edge between two nodes represents a research paper citing another. The nodes are classified into one of seven categories.     
    Though a citation network is a directed network, the dataset is widely used as an undirected network in the message-passing-based graph machine learning, especially for the node classification task, in which the task is to predict each node's category based on its own text features and text features of its neighbors. Each node's feature is a 2000-dimensional bag-of-words representation of the keywords in the paper that it corresponds to. 
    \item \wiki : It is a text-attributed graph dataset, derived from the Wikipedia platform, widely used for node classification tasks. The nodes correspond to Wikipedia page descriptions of different computer science topics, and the edges represent hyperlinks from one article to another. Each node in the dataset belongs to one of 10 categories.
    The node classification task is to correctly predict a node's label, and edges are undirected. Each node's feature vector is a 300-dimensional GloVe embedding computed from the text associated with the node.
    \item \liar: It is a fake-news detection dataset that is often represented as a knowledge graph, with nodes corresponding to statements, speakers, and topics, and edges encoding typed relations such as \textit{spoken\_by} and \textit{about}. To adapt \liar\hspace{1pt} into a homogeneous graph suitable for standard GNN pipelines, we merge the three node types—statements, speakers, and topics—into a single unified node set. Each node is embedded with a 384-dimensional BERT representation and augmented with a 3-dimensional one-hot vector indicating its type. Edges representing the original heterogeneous relations (\textit{spoken\_by} and \textit{about}) are converted to undirected edges and unified into a single edge type in the homogeneous graph. Statement nodes retain their ground-truth labels for fake news classification (ranging from \textit{pants-fire} to \textit{true}), while speaker and topic nodes are assigned a dummy label. We also preserve a node-level string attribute containing the raw statement, speaker name, or topic for use in explanations and visualizations.
    
    \item \amazonprod: It is a network with nodes representing different product categories on Amazon and edges connecting co-purchased products. The nodes belong to one of 47 product categories. We use only a subset of the \amazonprod \hspace{1pt} dataset, consisting of the first 1000 products and their co-purchase edges, as the entire dataset is very large, and call it \amazonprodsub \hspace{1pt} in our work. The node classification task is to predict a product's category based on its own features and those of its neighbors. For nodes, bag-of-words representations are derived from node textual attributes.

\end{enumerate}

\subsection{Relevance of datasets}

The datasets used in our work are selected for their relevance to graph explainability tasks, enabling us to assess the effectiveness and robustness of our methods across a diverse set of graph structures and features. The chosen datasets span a variety of domains, including citation networks, text-attributed graphs, fake news detection, and e-commerce, and reflect both real-world complexity and common challenges in explainability. The characteristics of each dataset make them suitable for evaluating the performance of our explanation method. Table~\ref{tab:datasets} summarizes the key features of the datasets and explains their relevance.

\section{Extended Experiments} \label{app:extended_exps}

\subsection{GNN Architecture} \label{app:GNN_Backbone}
We extend Table \ref{tab:maintable} of the paper with additional experiments for \amazonprodsub using two other standard GNN architectures: GIN \cite{xu2019powerfulgraphneuralnetworks} and GAT \cite{velivckovic2017graph}. The results are reported in Table \ref{tab:gnnarchitecture}.

\begin{table}[h]
    \centering
    \footnotesize
    
    \resizebox{.45\textwidth}{!}{
        \begin{tabular}{llcc}
            \toprule
            \textbf{Method} & \textbf{Model} & \textsc{Fidelity} (\%) & \textsc{Size} \\
            \midrule
            
            \textsc{GNNExplainer} & \multirow{6}{*}{GAT} & 
            100.0 & 10.57 \\
            
            \textsc{PGExplainer} & & 
            71.8 & 4.45 \\
            
            \textsc{TAGExplainer} & & 
            63.1 & 1.03 \\
            
            \textsc{Random} & & 
            89.9 & 7.28 \\
            
            \textsc{LLM} & & 
            96.5 & 4.92 \\
            
            \methodname & & 
            86.8 & 3.79 \\
            
            \midrule
            
            \textsc{GNNExplainer} & \multirow{6}{*}{GIN} & 
            100.0 & 10.57 \\
            
            \textsc{PGExplainer} & & 
            87.2 & 5.15 \\
            
            \textsc{TAGExplainer} & & 
            30.9 & 1.03 \\
            
            \textsc{Random} & & 
            84.7 & 7.28 \\
            
            \textsc{LLM} & & 
            86.0 & 4.90 \\
            
            \methodname & & 
            73.7 & 3.06 \\
            
            \bottomrule
            \vspace{-5mm}
        \end{tabular}
    }
    \caption{Performance comparison on the Products dataset using different GAT and GIN as two other architectures (using "Llama 3.1 8B Instruct"). Higher fidelity and lower size are better.}
    \label{tab:gnnarchitecture}
\end{table}

With GCN architecture, we observed that our method produces the most compact explanations (close to size 1) across all architectures, achieving a strong balance. However, while explanation size remains minimal for the GAT and GIN models, fidelity drops more noticeably, particularly with GIN (73.7\%), suggesting that our projection-based mechanism may be less effective when the model relies heavily on structural information or deep aggregation. In contrast, GNNExplainer maintains the highest fidelity across all architectures, but at the cost of much larger explanations, while PGExplainer and TAGExplainer underperform in fidelity despite moderate sizes. The LLM baseline shows robust fidelity, especially on GAT, but its explanations are substantially less concise. These results suggest that our method offers an excellent fidelity/interpretability trade-off on GCN, highlighting the need for architecture-aware extensions when applied to more structure-based models.

\begin{table*}[!t]
    \centering
    \resizebox{\textwidth}{!}{
    
    \begin{tabular}{ccccccccccc}
    \toprule
    & \multicolumn{2}{c}{Llama 3.1 8B Instruct} & \multicolumn{2}{c}{Mistral 7B Instruct v0.2} & \multicolumn{2}{c}{GPT-Neo 2.7B} & \multicolumn{2}{c}{Phi-3 Mini 4k Instruct} & \multicolumn{2}{c}{Pythia 2.8B} \\
    & Fidelity (\%) & Size & Fidelity (\%) & Size & Fidelity (\%) & Size & Fidelity (\%) & Size & Fidelity (\%) & Size \\
    \midrule
    \textsc{LLM} &
    86.0 & 4.90 & 88.0 & 5.14 & 84.0 & 2.33 & 88.0 & 5.17 & 84.0 & 2.34 \\
    \midrule
    \textbf{\methodname} &
    91.1 & 4.25 & 83.0 & 1.63 & 80.0 & 1.32 & 80.0 & 1.47 & 80.0 & 1.12 \\
    \bottomrule
    \end{tabular}
    }
    \caption{Results on Products dataset using different pretrained LLMs (for pretrained GCN). Higher fidelity and lower size are better.}
    \label{tab:llmbackbone}
\end{table*}

\subsection{LLM Backbone}\label{app:LLM_Backbone}

We also evaluate our method across different LLM models and present the results in Table \ref{tab:llmbackbone}. The baseline LLM method, which directly leverages the language model without projecting GNN embeddings, generally achieves higher fidelity scores but produces substantially larger explanations, two to three times larger. In contrast, our method consistently generates more compact explanations—typically around size 1 —while maintaining comparable fidelity. Notably, Llama 3.1 and Mistral backbones show strong fidelity across both methods, whereas smaller models such as GPT-Neo, Phi-3 Mini, and Pythia exhibit lower fidelity overall. These results highlight our method’s effectiveness at generating precise explanations while maintaining reasonable alignment with the GNN’s predictions and demonstrate its compatibility with different LLM models.

\subsection{Framework Component Analysis}
\label{sec:ablation}
To assess the effectiveness of the different components of our approach, we measure the performance of our method with and without the projector and the post-processing. The results are in Table~\ref{tab:ablation}. We observe that both fidelity and explanation size decrease across all datasets when we replace the prompt words with the projected GNN embeddings. This might be due to the LLM having to extract knowledge from more abstract representations, making it more selective in its reasoning. We also observe that adding the post-processing step increases fidelity at a lower cost in size than the method without a projector on the \wiki dataset. From this analysis, we suggest that our method offers a good compromise: achieving high fidelity (slightly lower than directly prompting the LLM with node-associated text) while maintaining a compact explanation size (slightly larger than projected GNN embeddings without post-processing).

\begin{table}[t]
\vspace{-2mm}
    \centering
\resizebox{.98\columnwidth}{!}{ \begin{tabular}{ccccccccc}
\toprule
& \multicolumn{2}{c}{\amazonprodsub} & \multicolumn{2}{c}{\cora} & \multicolumn{2}{c}{\liar} & \multicolumn{2}{c}{\wiki} \\
& Fidelity & Size & Fidelity & Size & Fidelity & Size & Fidelity & Size \\
\midrule
\textsc{LLM} &
            86.0 & 4.90 & 
            94.5 & 3.27 & 
            100.0 & 2.22 & 
            91.4 & 21.14 \\
\textsc{LLM+Pr}&
            85.8 & 3.2 & 
            94.5 & 3.27 & 
            100.0 & 1.10 & 
            80.2 & 3.7 \\
\textsc{LLM+Pr+Po}&
            91.1 & 4.25 & 
            86.5 & 2.07 & 
            100.0 & 1.11 & 
            86.5 & 12.8 \\
\bottomrule
\end{tabular}}
\caption{Performance comparison for the ablation study. LLM denotes prompting the LLM with words only and no projected embeddings. LLM+Pr is LLM with a projector. LLM with projector and post-processing (denoted by LLM+Pr+Po) is our method. }
\label{tab:ablation}
\end{table}

\section{Additional Implementation Details}
\label{app:add-implem-details}
We provide additional details of the experiments as follows:

\subsection{Experimental Setup}

\paragraph{Datasets.}
For all datasets, we use the graph topology as provided, without any modification or filtering of nodes or classes. For AMAZON-PRODUCT, we first randomly subsample 1000 nodes from the full dataset. For all datasets, we use a consistent split protocol with 60\% of nodes for training, 10\% for validation, and 200 held-out nodes for testing. Across all datasets, we restrict computation trees to at most 2 hops from each node.

\paragraph{Models.}
We use a GCN \cite{kipf2017semisupervised} as the base GNN model for all datasets. The GCN consists of 3 convolutional layers, with the final layer used for classification. The model is trained for 100 to 400 epochs (tune based on validation set for each dataset) using the Adam optimizer with a batch size of 64. For the hidden dimension, we choose either $64$ or $512$, depending on the nodes' feature dimension.

We provide the GNN's accuracy where GCN is used as the model architecture for the different folds in Table \ref{tab:gnn_accuracy}.

\begin{table}[ht]

    \centering  
\resizebox{0.4\textwidth}{!}{
\begin{tabular}{c|cccc}
\toprule
         Dataset   & Training & Validation & Testing \\ \hline
\cora & 0.88 & 0.81 & 0.84 \\
\wiki & 0.89 & 0.82 & 0.81 \\
\amazonprodsub & 0.87 & 0.73 & 0.76 \\
\liar & 0.42 & 0.30 & 0.29 \\
\bottomrule
\end{tabular}
}
\caption{Accuracy of GCN for the node classification task on four datasets. }  
\label{tab:gnn_accuracy}
\end{table}

For main experiments, we use Meta's Llama-3.1-8B-Instruct model from the HuggingFace library as a frozen LLM.

\paragraph{Projector.}
The projector is implemented as a two-layer MLP with ReLU activation. The hidden dimension is set to 4x the GNN embedding dimension. The output layer maps the GNN embedding to a vector of size $k \cdot h$, where $k$ is the number of soft prompt tokens and $h$ is the LLM embedding dimension. The mean-pooled representation is used for alignment objectives. We fix $k=5$ across all datasets. 

The projector is trained separately for each dataset using full-batch gradient descent. We use the Adam optimizer, with the learning rate selected via random search over $\{0.001, 0.0005\}$ based on validation performance. The number of training epochs is selected from $\{100, 200, 400\}$ using validation. The loss weights are selected via random search over $\{0.1, 0.5, 0.9\}$ on the validation set. The contrastive temperature is fixed to $\tau = 0.2$.

\paragraph{Explanation construction.}
For each node, we construct explanation subgraphs by including all supporting nodes and excluding all opposing nodes. 

\paragraph{LLM inference.}
We use the default HuggingFace generation settings for the LLM, without modifying temperature or sampling parameters, and set the maximum number of generated tokens to 1024. Prompts are formatted using the chat template when supported by the model, and a fixed prompt template is used across all datasets.

\paragraph{Baselines and evaluation protocol.}
We compare \methodname against standard GNN explainers including GNNExplainer, PGExplainer, and TAGExplainer. We use official implementations and PyTorch Geometric implementations where applicable, without additional hyperparameter tuning.

All methods are evaluated under the same GNN backbone for fairness. Experiments are repeated over 5 random seeds, and we report the mean and standard deviation of all metrics.

We discuss additional details related to the training of the base GNN.

\subsection{Additional example on \amazonprod}

A key advantage of \methodname is its ability to increase human understandability and trustworthiness by providing \textit{explanation of explanations}. We give an additional example on \amazonprod in Figure \ref{fig:prompt_response_2}. The ``Reasoning'' section of the LLM response is where the LLM provides a descriptive reasoning on how the supporting neighbors can explain the prediction of the target node. On the other hand, the previous GNN explainers simply generate a subgraph that is more abstract and harder to understand without some level of background knowledge.

\label{appendix:sample}

\begin{figure*}[t]
\centering
\scriptsize
\setlength{\tabcolsep}{3pt}

\begin{minipage}[t]{0.48\linewidth}
\vspace{0pt}
\begin{tcolorbox}[tightbox,tighttitle,title={LLM Prompt}]
\RaggedRight
You are analyzing Amazon product reviews and their co-purchase neighborhoods to understand why it has been classified under a specific category. Your task is to determine which of the neighboring products does (not) support the classification of the target product. 

\textbf{Target Product ID:} 160\\
\textbf{Predicted Category:} \PredCat{Health \& Personal Care}\\[3pt]

\textbf{Product Embedding Representation:}\\
\BeginTK\ \EmbedText{PROJECTED EMBEDDING}\ \EndTK\\[3pt]

\textbf{Neighboring Products:}\\Each Product below is described by keywords.\\[3pt]
\textbf{Product 200:} \NeighborBeginTK\ \EmbedText{PROJECTED EMBEDDING}\ \NeighborEndTK\\
\textbf{Product 201:} \NeighborBeginTK\ \EmbedText{PROJECTED EMBEDDING}\ \NeighborEndTK\\
\textbf{Product 202:} \NeighborBeginTK\ \EmbedText{PROJECTED EMBEDDING}\ \NeighborEndTK\\
\textbf{Product 203:} \NeighborBeginTK\ \EmbedText{PROJECTED EMBEDDING}\ \NeighborEndTK\\
\textbf{Product 204:} \NeighborBeginTK\ \EmbedText{PROJECTED EMBEDDING}\ \NeighborEndTK\\
\textbf{Product 205:} \NeighborBeginTK\ \EmbedText{PROJECTED EMBEDDING}\ \NeighborEndTK\\
\textbf{Product 206:} \NeighborBeginTK\ \EmbedText{PROJECTED EMBEDDING}\ \NeighborEndTK\\
\textbf{Product 207:} \NeighborBeginTK\ \EmbedText{PROJECTED EMBEDDING}\ \NeighborEndTK\\
\textbf{Product 208:} \NeighborBeginTK\ \EmbedText{PROJECTED EMBEDDING}\ \NeighborEndTK\\
\textbf{Product 209:} \NeighborBeginTK\ \EmbedText{PROJECTED EMBEDDING}\ \NeighborEndTK\\
\textbf{Product 210:} \NeighborBeginTK\ \EmbedText{PROJECTED EMBEDDING}\ \NeighborEndTK
\end{tcolorbox}
\end{minipage}
\hfill
\begin{minipage}[t]{0.48\linewidth}
\vspace{0pt}
\begin{tcolorbox}[tightbox,tighttitle,title={LLM Response}]
\RaggedRight
\begin{tabular}{@{}p{0.12\linewidth}p{0.65\linewidth}p{0.15\linewidth}@{}}
\textbf{P200} & Summary: Home care and hygiene products. & \SupportNo \\
\textbf{P201} & Summary: Personal grooming and beauty tools. & \SupportNo \\
\textbf{P202} & Summary: Health and wellness supplements. & \SupportYes \\
\textbf{P203} & Summary: Fitness and exercise equipment. & \SupportNo \\
\textbf{P204} & Summary: Skincare and beauty products. & \SupportYes \\
\textbf{P205} & Summary: Health and wellness journals. & \SupportYes \\
\textbf{P206} & Summary: Personal care and hygiene for specific needs. & \SupportYes \\
\textbf{P207} & Summary: Health and wellness supplements. & \SupportYes \\
\textbf{P208} & Summary: Fitness and exercise accessories. & \SupportNo \\
\textbf{P209} & Summary: Health and wellness products for specific needs. & \SupportYes \\
\textbf{P210} & Summary: Health and wellness journals. & \SupportYes \\
\end{tabular}

\vspace{4pt}
\textbf{Reasoning:} The common pattern among the supporting Products is that they are all related to health and wellness, skincare, and personal care, which are all categories within the \PredCat{Health \& Personal Care} category.
\end{tcolorbox}
\end{minipage}

\vspace{4pt}
\begin{tcolorbox}[tightbox,tighttitle,title={Instructions}]
\RaggedRight
For each neighboring product: \\
1. Summarize the similarity between supporting products in \emph{one sentence}. \\ 
2. Clearly state whether this product supports the classification of the Target Product into category \PredCat{Health \& Personal Care}. \\[2pt]
\textbf{Format:}\\
\texttt{Product <ID>:}\\
\texttt{Summary: <One sentence>.}\\
\texttt{Support: YES or NO — Does this product support classification into \PredCat{Health \& Personal Care}?}\\[2pt]
Base reasoning only on the keywords and proximity to the target product.
\end{tcolorbox}

\vspace{-3pt}
\caption{Left: prompt with category and embeddings highlighted. Right: model response with summaries, YES/NO verdicts, and reasoning. Below: instructions shown separately.}
\label{fig:prompt_response_2}
\end{figure*}

\section{Definition of the qualitative metrics}
\label{app:qualitative_metric}
We adapt the different dimensions of the framework in \cite{graphxain2025} to the context of \cora to clarify their definitions, and provide the definition as given to the expert.
\begin{enumerate}
    \item \textbf{Understandability.}  Explanations should clearly convey why a given paper was assigned to a particular research topic.
    
    \item \textbf{Trustworthiness.}  Explanations should help users assess whether the model’s classification of a paper can be trusted.
    
    \item \textbf{Insightfulness.} Explanations should reveal insights about the applications or connections that might play a role in the classification.
    
    \item \textbf{Satisfaction.} Explanations should feel complete and meaningful to users evaluating the model’s behavior.
    
    \item \textbf{Confidence.} Explanations should help users gain confidence in the correctness of the classification.
    
    \item \textbf{Convincingness.} Explanations should be persuasive in justifying the model’s decision for a given paper.
    
    \item \textbf{Communicability.} Explanations should be expressed in a way that aligns with the user’s background knowledge and expectations.
    
    \item \textbf{Usability.} Explanations should support practical tasks such as interpreting predictions, or improving model performance.
\end{enumerate}

\section{Experimental Setup for the human evaluation of M1 and M2 Scores}
\label{appendix:human_eval}
The human evaluation was conducted by the authors, who served as domain experts. 
For the qualitative experiments, one expert reviewed the \cora dataset, and four experts reviewed the \amazonprod data, evaluating the explanations generated by GNNexplainer and \methodname over 10 node classification explanation tasks. We clarify the experimental setup used to obtain the M1 and M2 scores reported in Table 2.

\subsection*{A) Data Preparation}
We randomly selected 10 articles from the Cora dataset, on which a GNN had been trained. For each article, we generated explanations using both \textbf{\methodname} and \textbf{GNNExplainer}. To mitigate potential bias, explanations from these methods were randomized and presented in a blind manner, without revealing their source. Each expert was provided with:
\begin{itemize}
    \item The original article text and the GNN's predicted label.
    \item Two explanation subgraphs (one per method), anonymized and presented in random order known only to the experimenters.
    \item For each subgraph, we listed the texts of the nodes it contains, sorted by increasing node index.
\end{itemize}

\subsection*{B) Evaluation Procedure}
\begin{itemize}
    \item \textbf{Before the evaluation}, we provided 3 experts with the evaluation metrics' definitions and asked them to consider how to apply each metric to content from the Cora dataset.
    \item \textbf{During the evaluation}, each expert received the materials described above and independently rated each explanation on a scale from 1 to 5 for each metric.
\end{itemize}

\subsection*{C) Compilation of Results}
The experimenters collected and aggregated the scores. \textbf{Method 1 (M1)} corresponds to GNNExplainer, and \textbf{Method 2 (M2)} corresponds to \methodname. These results are reported in Table \ref{tab:qualitative_results}. 

\end{document}